\documentclass[10pt,twocolumn,letterpaper]{article}

\usepackage[pagenumbers]{cvpr} 

\usepackage{graphicx}
\usepackage{amsmath}
\usepackage{amssymb}
\usepackage{booktabs}
\usepackage{setspace}
\usepackage{algorithmic}
\usepackage{algorithm}
\usepackage{bbm}

\usepackage[pagebackref,breaklinks,colorlinks]{hyperref}

\usepackage[capitalize]{cleveref}
\crefname{section}{Sec.}{Secs.}
\Crefname{section}{Section}{Sections}
\Crefname{table}{Table}{Tables}
\crefname{table}{Tab.}{Tabs.}

\begin{document}


\title{HCSC: Hierarchical Contrastive Selective Coding}

\author{
	Yuanfan Guo\textsuperscript{\rm 1$\,*,\dagger$} \quad
	Minghao Xu\textsuperscript{\rm 2,3$\,*,\ddagger$} \quad
	Jiawen Li\textsuperscript{\rm 4} \quad
	Bingbing Ni\textsuperscript{\rm 1}\vspace{0.5mm}\\
	Xuanyu Zhu\textsuperscript{\rm 4} \quad
	Zhenbang Sun\textsuperscript{\rm 4} \quad
	Yi Xu\textsuperscript{\rm 1$\,\S$}\vspace{0.5mm}\\
	\textsuperscript{\rm *}{\small equal contribution} \quad
	\textsuperscript{\rm $\dagger$}{\small technique lead} \quad
	\textsuperscript{\rm $\ddagger$}{\small project lead} \quad
	\textsuperscript{\rm $\S$}{\small corresponding author}\vspace{0.5mm}\\
	\textsuperscript{\rm 1}MoE Key Lab of Artificial Intelligence, AI Institute, Shanghai Jiao Tong University \\
	\textsuperscript{\rm 2}Mila - Qu\'{e}bec AI Institute \;
	\textsuperscript{\rm 3}University of Montr\'{e}al \;
	\textsuperscript{\rm 4}ByteDance\vspace{0.5mm}\\
	{\small contacts: Yuanfan Guo $<$gyfastas@sjtu.edu.cn$>$, Minghao Xu $<$minghao.xu@umontreal.ca$>$, Yi Xu $<$xuyi@sjtu.edu.cn$>$}\vspace{1mm}
}

\maketitle

\begin{abstract}

Hierarchical semantic structures naturally exist in an image dataset, in which several semantically relevant image clusters can be further integrated into a larger cluster with coarser-grained semantics. Capturing such structures with image representations can greatly benefit the semantic understanding on various downstream tasks. Existing contrastive representation learning methods lack such an important model capability. In addition, the negative pairs used in these methods are not guaranteed to be semantically distinct, which could further hamper the structural correctness of learned image representations. To tackle these limitations, we propose a novel contrastive learning framework called \textbf{H}ierarchical \textbf{C}ontrastive \textbf{S}elective \textbf{C}oding (HCSC). In this framework, a set of hierarchical prototypes are constructed and also dynamically updated to represent the hierarchical semantic structures underlying the data in the latent space. To make image representations better fit such semantic structures, we employ and further improve conventional instance-wise and prototypical contrastive learning via an elaborate pair selection scheme. This scheme seeks to select more diverse positive pairs with similar semantics and more precise negative pairs with truly distinct semantics. On extensive downstream tasks, we verify the superior performance of HCSC over state-of-the-art contrastive methods, and the effectiveness of major model components is proved by plentiful analytical studies. We build a comprehensive model zoo in Sec.~\ref{supp_sec4}. Our source code and model weights are available at \url{https://github.com/hirl-team/HCSC}.

\end{abstract}



\section{Introduction} 
\label{sec1}

In the past few years, self-supervised image representation learning has witnessed great progresses, in which the traditional methods based on solving informative puzzles~\cite{doersch2015unsupervised,noroozi2016unsupervised,zhang2016colorful,noroozi2017representation,gidaris2018unsupervised} are obviously surpassed by contrastive learning approaches~\cite{CPC,MOCO,SimCLR,moco_v2,SwAV}. These contrastive methods succeed in deriving useful and interpretable feature representations for various downstream tasks. In particular, under the standard linear evaluation protocol~\cite{NPID}, they have achieved inspiring results that approach fully-supervised learning. 

Existing contrastive approaches can be mainly classified into two categories, \emph{instance-wise contrastive learning}~\cite{CPC,MOCO,SimCLR} and \emph{prototypical contrastive learning}~\cite{SwAV,PCL}. Instance-wise contrast seeks to map similar instances nearby in the latent space while mapping dissimilar ones far apart, which guarantees reasonable local structures among different image representations. Prototypical contrast aims to derive compact image representations gathering around corresponding cluster centers, which captures some basic semantic structures that can be represented by a single hierarchy of clusters. 



\begin{figure}[tb]
\centering
    \includegraphics[width=0.95\linewidth]{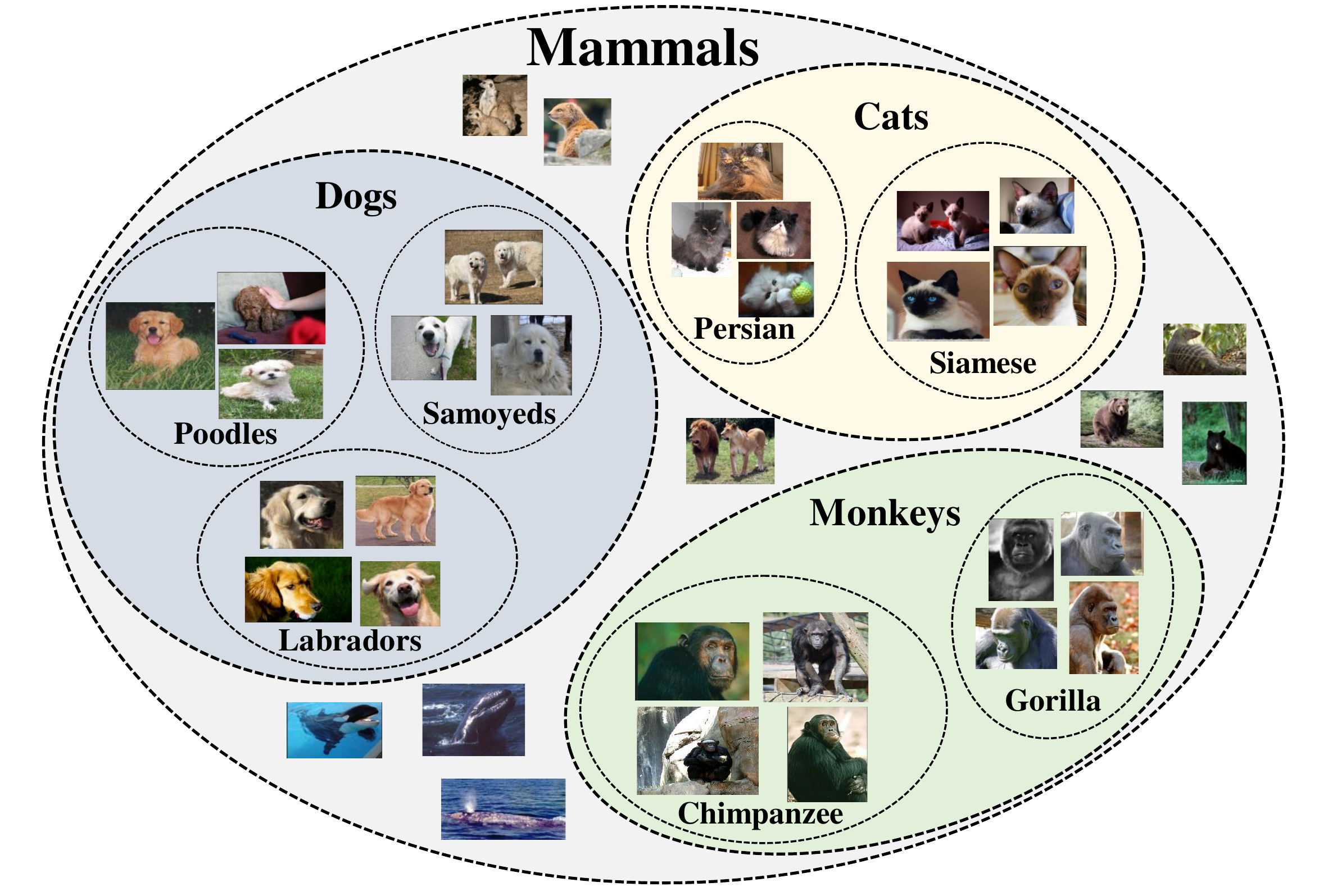}
    \vspace{-2.7mm}
    \caption{An image dataset always contains multiple semantic hierarchies, \emph{e.g.} ``mammals $\rightarrow$ dogs $\rightarrow$ Labradors'' in the order from coarse-grained semantics to fine-grained semantics.}
    \vspace{-3.5mm}
    \label{fig:motivation}
\end{figure}


However, these approaches lag in representation power when modeling \textbf{a large-scale image dataset which could always possess multiple semantic hierarchies}. For example, in an extensive species database, the cluster of dogs summarizes the common characters of Labradors, Poodles, Samoyeds, \emph{etc.} and should be placed on a higher hierarchy; similarly, dogs together with cats, monkeys, whales, \emph{etc.} are further summarized by an even higher-level cluster, mammals (see Fig.~\ref{fig:motivation} for a more intuitive illustration). Learning image representations that embrace such hierarchical semantic structures can greatly benefit the semantic understanding on various downstream tasks. How to achieve this by contrastive learning is still an open problem.

In addition, existing contrastive methods commonly construct negative pairs by exhaustive sampling from some noise distribution, and all the sampled negative pairs are used without selection. There is no guarantee that the negative pairs obtained in this way own truly distinct semantics. Therefore, some samples with similar semantics may be wrongly embedded far apart by these methods, which hampers the quality of learned image representations.


To tackle the limitations above, we propose a novel contrastive learning framework called \textbf{H}ierarchical \textbf{C}ontrastive \textbf{S}elective \textbf{C}oding (\emph{HCSC}). In this framework, we propose to capture the hierarchical semantic structures underlying the data with \emph{hierarchical prototypes}, a set of tree-structured representative embeddings in the latent space. Along the training process, these prototypes are dynamically updated to fit the current image representations. Under the guidance of such hierarchical semantic structures, we seek to improve both instance-wise and prototypical contrastive learning by \emph{selecting high-quality positive and negative pairs that are semantically correct}. Specifically, for each query sample, we search for its most similar prototype on each semantic hierarchy to build more abundant positive pairs. Moreover, 
for each candidate of negative pair, we conduct a Bernoulli sampling to keep/discard it if the semantic correlation of the pair is low/high. By using these selected pairs for~instance-wise and prototypical contrast, the semantic constraints from hierarchical prototypes can be embedded into the objective of representation learning.


\vspace{0.5mm}
We summarize the contributions of this work as follows:
\vspace{-4.5mm}
\begin{itemize}
    \item We novelly propose to represent the hierarchical semantic structures of image representations by dynamically maintaining hierarchical prototypes.
    \item We propose a novel contrastive learning framework, Hierarchical Contrastive Selective Coding (HCSC), which improves conventional instance-wise and prototypical contrastive learning by selecting semantically correct positive and negative pairs. 
    \item Our HCSC approach consistently achieves superior performance over state-of-the-art contrastive learning algorithms on various downstream tasks. Also, the effectiveness of key model components are verified by extensive ablation and visualization analysis. 
\end{itemize}






\section{Related Work}


\subsection{Self-supervised Representation Learning}

\textbf{Solving Pretext Puzzles.} Most early works for self-supervised image representation learning aim to solve pretext puzzles, \emph{e.g.} counting objects~\cite{noroozi2017representation}, solving the jigsaw puzzle~\cite{noroozi2016unsupervised}, recovering colors from gray-scale images~\cite{zhang2016colorful,larsson2016learning}, rotation prediction~\cite{gidaris2018unsupervised}, \emph{etc.} These pretext tasks are not guaranteed to derive discriminative feature representations for different downstream tasks.

\textbf{Instance-wise Contrastive Learning.} The instance-wise contrast approaches seek to embed similar instances nearby in the latent space while embed dissimilar ones far apart. 
Standard instance-wise contrastive methods~\cite{NPID,SimCLR,MOCO} achieve this goal by maximizing the mutual information between correlated instances, \emph{i.e.} optimizing with an InfoNCE loss~\cite{CPC}. Recent works improve such a standard scheme by employing inter-instance positive pairs~\cite{NNCLR}, introducing stronger augmentation functions~\cite{CLSA} or designing predictive pretext tasks that are free from negative sampling~\cite{BYOL,SimSiam,BarlowTwins}. However, these methods are not aware of the global semantics underlying the whole dataset. 

\textbf{Prototypical Contrastive Learning.} Another series of contrastive approaches seek to explicitly exploit semantic structures by utilizing the prototype representations of image clusters.
They either contrast between correlated and uncorrelated prototype pairs~\cite{SwAV,contrastive_clustering} or between associated and unassociated instance-prototype pairs~\cite{PCL,CLD}, which derives more semantically compact image representations. However, all these methods represent semantic clusters at a single hierarchy, which neglects the important fact that an image dataset naturally possesses hierarchical semantics. 

\emph{Improvements over existing works.} In this work, we novelly propose to construct and maintain hierarchical semantic structures of image representations, which aligns with a recent effort~\cite{xu2021self} on learning hierarchical molecular representations. In addition, we improves both instance-wise and prototypical contrastive learning by selecting high-quality positive and negative pairs under precise semantic guidance.

\subsection{Deep Clustering}

Our work also relates to deep clustering, \emph{i.e.} performing clustering in a low-dimensional embedding space. A line of research~\cite{MitigatingClustering,IIC} aims at more accurate clustering by leveraging a clustering-favored latent space. Another line of research~\cite{xie2016unsupervised,yang2016joint,liao2016learning,yang2017towards,DeepCluster,deepercluster,SELA,PCL} jointly learns clustering assignments and image representations. Most of these methods~\cite{xie2016unsupervised,yang2016joint,DeepCluster} utilize some standard clustering algorithm, like K-means~\cite{k-means} or agglomerative clustering~\cite{agglomerative}, to establish a single hierarchy of semantic clusters, which is not sufficient to represent the semantic hierarchies underlying a set of images.


\emph{Improvements over existing works.} Compared to most existing works that learn a single semantic hierarchy, in this work, we seek to learn image representations with multiple semantic hierarchies. DeeperCluster~\cite{deepercluster} made an attempt on this direction by predicting clustering assignments hierarchically. By comparison, our method leverages contrastive selective coding to discover the semantic hierarchies of the data more accurately.

\section{Problem Definition and Preliminaries} \label{sec3}


\subsection{Problem Definition} \label{sec3_1}

Given a set $X = \{ x_1, x_2, \cdots, x_N \}$ of $N$ unlabeled images, we aim to learn a low-dimensional vector $z_n \in \mathbb{R}^{\delta}$ for each $x_n \in X$. Besides image representations, we together maintain a set of hierarchical prototypes $C = \{ \{ c^l_i \}_{i=1}^{M_l} \}_{l=1}^{L}$ to characterize the hierarchical semantic structures underlying the data, where $L$ stands for the number of semantic hierarchies, and $M_l$ is the number of prototypes in the $l$-th hierarchy. Each prototype $c^l_i \in C$ is also represented as a $\delta$-dimensional vector. Following the self-supervised learning protocol, image representations $Z = \{z_1, z_2, \cdots, z_N\}$ and hierarchical prototypes $C$ are both learned or maintained under the guidance of the data itself.


\subsection{Preliminaries} \label{sec3_2}

\textbf{Instance-wise contrastive learning.} To achieve the goal of self-supervised representation learning, a widely-used way is to contrast between a positive instance pair with some negative pairs. Specifically, given the representations $(z, z')$ of a pair of correlated instances, a standard InfoNCE loss~\cite{CPC} is defined to maximize the similarity between this positive pair and minimize the similarities between some randomly sampled negative pairs:
\begin{equation} \label{eq1}
\mathcal{L}_{\mathrm{InfoNCE}}(z, z', \mathcal{N}, \tau) = - \log \frac{ \exp(z \cdot z' / \tau)}{\sum_{z_j \in \{z'\} \cup \mathcal{N}} \exp(z \cdot z_j / \tau)} ,
\end{equation}
where $\mathcal{N}$ is a set of negative samples for $z$, and $\tau$ denotes a temperature parameter. 


\textbf{Prototypical contrastive learning.} In this contrastive representation learning manner, each positive pair consists of an instance and its associated semantic prototype, and negative pairs are formed by pairing instances with irrelevant semantic prototypes. In this way of pair construction, given a positive pair $(z, c)$, the ProtoNCE loss~\cite{PCL} is defined based on the same rationale of InfoNCE loss:
\begin{equation} \label{eq2}
\small
\mathcal{L}_{\mathrm{ProtoNCE}}(z, c, \mathcal{N}_c, \{\tau_c\}) = - \log \frac{ \exp(z \cdot c / \tau_c)}{\sum_{c_j \in \{c\} \cup \mathcal{N}_c} \exp(z \cdot c_j / \tau_{c_j})} ,
\end{equation}
where $\mathcal{N}_c$ stands for a set of negative prototypes for instance representation $z$, and $\tau_c$ is a prototype-specific temperature parameter which can be adaptively determined by some clustering statistics. 


\section{Method} \label{sec4}


\subsection{Motivation and Overview} \label{sec4_1}

Many recent efforts have been paid on learning informative visual representations with contrastive methods. Most of these works~\cite{CPC,MOCO,SimCLR} focused on probing instance-wise relationships, and several works~\cite{SwAV,PCL} further attempted to discover the semantic structures within the data during contrastive learning. These approaches represented semantic structures with a group or several independent groups of cluster centers, which cannot represent the semantic hierarchies that naturally exist in an image dataset. For example, the images of Labradors, Poodles, Samoyeds, \emph{etc.} form a higher-level cluster of dogs, and dogs together with cats, monkeys, whales, \emph{etc.} can be further clustered as mammals. Capturing such hierarchical semantics promises the stronger representation power of an image encoder, while it has not been achieved by previous methods. Under the context of contrastive learning in particular, these semantic structures can give helpful guidance to select positive pairs with similar semantics and negative pairs with distinct semantics, which is also less explored by existing works.

\begin{figure*}[htb]
\centering
    \includegraphics[width=1.0\linewidth]{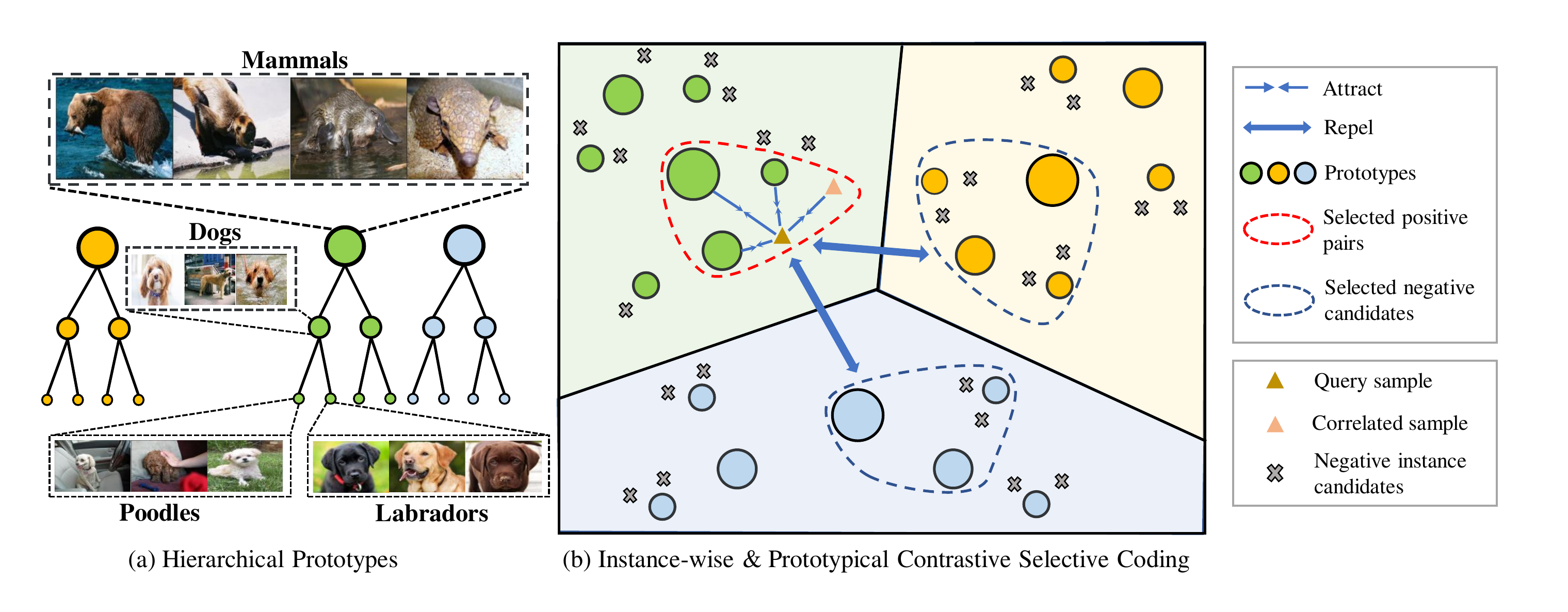}
    \vspace{-9mm}
    \caption{\textbf{Illustration of HCSC framework.} (a) In the latent space, a set of \emph{hierarchical prototypes} are used to represent the hierarchical semantic structures underlying an image dataset. (b) \emph{Instance-wise and prototypical contrastive selective coding} select semantically correct positive and negative pairs for contrastive learning, which is guided by the semantic information from hierarchical prototypes.}
    \label{fig:framework}
\vspace{-2mm}
\end{figure*}

Motivated by these limitations, we propose a novel contrastive learning framework called \textbf{H}ierarchical \textbf{C}ontrastive \textbf{S}elective \textbf{C}oding (\emph{HCSC}). In a nutshell, we represent the semantic structures of the data with \emph{hierarchical prototypes} and dynamically update these prototypes along the training process (Sec.~\ref{sec4_2}). Based on such hierarchical semantic representations, we seek to boost conventional instance-wise and prototypical contrastive learning by selecting better positive and negative pairs that fit the semantic structures (Secs.~\ref{sec4_3} and \ref{sec4_4}), and our overall objective combines both learning manners (Sec.~\ref{sec4_5}). The graphical summary of our approach is shown in Fig.~\ref{fig:framework}. 


\begin{algorithm}[tb] 
   \caption{Hierarchical K-means.}
   \label{alg:clustering}
\begin{spacing}{1.05}
\begin{algorithmic}
   \STATE {\bfseries Input:} Image representations $Z$, \#~semantic hierarchies $L$, \#~prototypes at the $l$-th hierarchy $M_l$.
   \STATE {\bfseries Output:} Hierarchical prototypes $C = \{ \{ c^l_i \}_{i=1}^{M_l} \}_{l=1}^{L}$, the undirected edges $E$ between different prototypes.
   \STATE $\{ c^1_i \}_{i=1}^{M_1} \gets \textrm{K-means}(Z)$.
   \FOR{$l=2$ {\bfseries to} $L$}
   \STATE $\{ c^l_i \}_{i=1}^{M_l} \gets \textrm{K-means}\big( \{ c^{l-1}_i \}_{i=1}^{M_{l-1}} \big)$.
   \FOR{$i=1$ {\bfseries to} $M_{l-1}$}
   \STATE $E \gets E \cup \big\{ \big( c^{l-1}_i, \textrm{Parent}(c^{l-1}_i) \big) \big\}$.
   \ENDFOR
   \ENDFOR
\end{algorithmic}
\end{spacing}
\end{algorithm}


\subsection{Hierarchical Semantic Representation} \label{sec4_2}

The core of HCSC framework is to construct and maintain the hierarchical semantic structures of the data in the latent space. Compared to previous methods~\cite{SwAV,PCL} that can only capture a single semantic hierarchy or several independent hierarchies, our approach is able to capture the hierarchical relations among various semantic clusters. We achieve this by a simple yet effective \emph{hierarchical K-means} algorithm that performs in a bottom-up manner. 

The detailed algorithm is summarized in Alg.~\ref{alg:clustering}. First, the feature representations of all images in the dataset are extracted by the image encoder, and K-means clustering is applied upon these image representations to obtain the prototypes of the first hierarchy. After that, the prototypes of each higher hierarchy are derived by iteratively applying K-means clustering to the prototypes of hierarchy below. To construct the hierarchical semantic structure, we further connect each prototype $c^{l-1}_i$ with its parent prototype $c^l_j$ at the higher hierarchy, in which $c^{l-1}_i$ is assigned to $c^l_j$ during K-means clustering. All such connections form an undirected edge set $E$. In this way, the hierarchical prototypes are structured as a set of trees (Fig.~\ref{fig:framework}(a)). 
In this algorithm, the number $L$ of semantic hierarchies and the number $M_l$ of prototypes at the $l$-th hierarchy are specified in Sec.~\ref{sec5_1}, and their sensitivities are studied in Sec.~\ref{sec6_2}.

Since image representations are updated along the training process, a maintenance scheme on hierarchical prototypes is also required to ensure that they are representative embeddings in the latent space. In our implementation, to balance between precision and efficiency, we conduct the hierarchical K-means algorithm before the start of each epoch to update hierarchical prototypes according to the current image representations. We analyze the time complexity of this scheme in Sec.~\ref{supp_sec5}. Upon such hierarchical prototypes, we seek to promote instance-wise and prototypical contrastive learning. 


\subsection{Instance-wise Contrastive Selective Coding} \label{sec4_3}

The gist of instance-wise contrastive learning is to embed similar instances nearby in the latent space while embed those dissimilar ones far apart. It is easy to obtain a similar (\emph{i.e.} positive) instance pair via data augmentation,
while the definition of dissimilar (\emph{i.e.} negative) instance pair is nontrivial. Previous methods derived negative samples by sampling uniformly over the dataset~\cite{MOCO,SimCLR,moco_v2} or sampling from a debiased data distribution~\cite{debiased_contrast}. However, they cannot guarantee that the produced negative samples own exactly distinct semantics relative to the query sample. Such a defect hampers instance-wise contrastive learning, in which those semantically relevant positive candidates could be wrongly expelled from the query sample in the latent space, and the semantic structure is thus broken to some extent. To overcome this drawback, we aim to select more precise negative samples that own truly irrelevant semantics with the query.

For a specific query image $x$, instead of contrasting it indiscriminately with all negative candidates $\mathcal{N}$ in a queue~\cite{MOCO,moco_v2}, we select truly negative samples for contrasting by performing Bernoulli sampling on each negative candidate. Intuitively, in such a sampling process, we would like to eliminate those candidates sharing highly similar semantics with the query, while keep the ones that are less semantically relevant to the query. To achieve this goal, we first define a similarity measure between an image and a semantic cluster. Following PCL~\cite{PCL}, for a semantic cluster represented by prototype $c \in C$, we define the semantic similarity between image representation $z$ and this cluster using a cluster-specific dot product:
\begin{equation} \label{eq3}
s(z, c) = \frac{z \cdot c}{\tau_c}, \quad \tau_c = \frac{\sum_{z_i \in Z_c} ||z_i - c||_2} {|Z_c| \log (|Z_c| + \epsilon )},
\end{equation}
where $Z_c$ consists of the representations of the images assigned to cluster $c$ 
(details of constructing $Z_c$ are stated in Sec.~\ref{sec5_1}), and $\epsilon$ is a smooth parameter balancing the scale of temperature $\tau_c$ among different clusters. 

On such basis, we conduct negative sample selection at each semantic hierarchy. At the $l$-th hierarchy, we denote the cluster that owns highest semantic similarity with the query image as $c^{l}(z) = \mathop{\arg\max}_{c \in \{ c^l_i \}_{i=1}^{M_l}} s(z, c)$ ($z$ is the representation of query). For a negative candidate $z_j \in \mathcal{N}$, we are more likely to select it if its similarity with $c^{l}(z)$ is less prominent compared its similarities with other clusters at the same hierarchy. Based on such an intuition, we define the selected probability of $z_j$ with the following formula:
\vspace{-1mm}
\begin{equation} \label{eq4}
    p^l_{\mathrm{select}}(z_j; z) = 1 - \frac{\exp \big[ s(z_j, c^{l}(z)) \big]}{\sum_{i=1}^{M_l} \exp \big[ s(z_j, c^l_i) \big]} .
\end{equation}
After that, a Bernoulli sampling is performed on each negative candidate to derive more precise negative samples for the specific query:
\vspace{-1mm}
\begin{equation} \label{eq5}
    \mathcal{N}^l_{\mathrm{select}} (z) = \big\{ \mathcal{B} \big(z_j; p^l_{\mathrm{select}}(z_j; z) \big) | z_j \in \mathcal{N} \big\} ,
\end{equation}
where $\mathcal{B}(z; p)$ denotes a Bernoulli trial of accepting $z$ with probability $p$. Such a selection scheme is performed on all $L$ semantic hierarchies, which diversifies the composition of negative samples.
In this way, $L$ negative sample sets, \emph{i.e.} $\{ \mathcal{N}^l_{\mathrm{select}} (z) \}_{l=1}^{L}$, are produced for the query. We explicitly evaluate this negative sample selection scheme in Sec.~\ref{supp_sec6}.

By using these refined negative samples, we define the objective function of instance-wise contrastive selective coding (ICSC) as below:
\vspace{-1mm}
\begin{equation} \label{eq6}
\mathcal{L}_{\mathrm{ICSC}} = \mathbb{E}_{x \sim p_d} \big[ \frac{1}{L} \sum_{l=1}^{L} \mathcal{L}_{\mathrm{InfoNCE}}(z, z', \mathcal{N}^l_{\mathrm{select}} (z), \tau) \big] ,
\end{equation}
where $p_d$ denotes the data distribution, $z$ and $z'$ are the representations of $x$ and an augmented view of $x$, respectively. We fix the temperature parameter $\tau$ as $0.2$ for instance-wise contrast following~\cite{moco_v2}.


\subsection{Prototypical Contrastive Selective Coding} \label{sec4_4}

Prototypical contrastive learning aims to derive compact image representations in the latent space, in which each image is closely embedded surrounding its associated cluster center.
Given a query sample, previous works~\cite{SwAV,PCL} compare it with a single pool of prototypes, and they regard its most similar prototype as its positive companion and all other prototypes as negative samples. However, this scheme neglects the semantic correlation among different clusters, and it could excessively penalize on some semantically relevant clusters. For example, given an image of Labradors, it should not be strongly expelled from the image clusters of other kinds of dogs, considering the semantic correlation among these clusters. To mitigate such over-penalization, we seek to select more exact negative clusters that are semantically distant from the query.

Given a query image represented by embedding vector $z$, we first retrieve its most similar prototype at the $l$-th hierarchy, \emph{i.e.} $c^{l}(z) = \mathop{\arg\max}_{c \in \{ c^l_i \}_{i=1}^{M_l}} s(z, c)$, using the same similarity measure as in instance-wise contrast (Eq.~\eqref{eq3}). We regard $(z, c^{l}(z))$ as a positive pair, and the rest prototypes at that hierarchy serve as the candidates of negative clusters, denoted as $\mathcal{N}^l$. Since some of these negative candidates might own similar semantics as the positive cluster $c^{l}(z)$, 
it is unreasonable to embed the query distantly from these semantically relevant clusters. Therefore, we aim to perform prototypical contrast with the candidates that are more semantically distant from $c^{l}(z)$. To attain this goal, we take advantage of the semantic structures captured by hierarchical prototypes,
and a Bernoulli sampling is performed on each negative candidate under such guidance.

Concretely, according to the edges of hierarchical prototypes, we first identify the parent node of $c^{l}(z)$ in its corresponding tree structure, denoted as $\textrm{Parent}(c^{l}(z))$. This parent cluster summarizes the common semantics shared by its child clusters at the $l$-th hierarchy, \emph{e.g.} the general characters of different types of dogs. The negative clusters used for prototypical contrast are desired to share a semantic correlation with $\textrm{Parent}(c^{l}(z))$ as low as possible. Therefore, we are more likely to select a negative candidate $c_j \in \mathcal{N}^l$ if its similarity with $\textrm{Parent}(c^{l}(z))$ does not dominate its similarities with other clusters at the $(l+1)$-th hierarchy, which derives the selected probability of $c_j$ as below:
\begin{equation} \label{eq7}
p^l_{\mathrm{select}} (c_j; c^{l}(z)) = 1 - \frac{\exp \big[ s \big( c_j, \textrm{Parent}(c^{l}(z)) \big) \big]}{\sum_{i=1}^{M_{l+1}} \exp \big[ s(c_j, c^{l+1}_i) \big]} ,
\end{equation}
where the similarity measure $s(\cdot,\cdot)$ follows Eq.~\eqref{eq3}. We then conduct a Bernoulli sampling on each negative candidate to select exactly negative samples for prototypical contrast: 
\begin{equation} \label{eq8}
    \mathcal{N}^l_{\mathrm{select}} (c^{l}(z)) = \big\{ \mathcal{B} \big(c_j; p^l_{\mathrm{select}}(c_j; c^{l}(z)) \big) | c_j \in \mathcal{N}^l \big\} .
\end{equation}
Note that, we use all the negative candidates $\mathcal{N}^L$ at the top hierarchy without selection, since the semantic clusters at the top hierarchy share little semantic correlation and can thus be safely deemed as negative samples of each other. 

By contrasting the positive pair $(z, c^{l}(z))$ with these selected negative samples, the objective function of prototypical contrastive selective coding (PCSC) is defined as below:
\begin{equation} \label{eq9}
\small
\mathcal{L}_{\mathrm{PCSC}} = \mathbb{E}_{x \sim p_d} \big[ \frac{1}{L} \sum_{l=1}^{L} \mathcal{L}_{\mathrm{ProtoNCE}}\big( z, c^{l}(z), \mathcal{N}^l_{\mathrm{select}} (c^{l}(z)), \{\tau_c\} \big) \big] ,
\end{equation}
where $p_d$ denotes the data distribution, $z$ is the representation of image $x$, and we have $\mathcal{N}^L_{\mathrm{select}} (c^{L}(z)) = \mathcal{N}^L$ at the top hierarchy. We use a cluster-specific temperature parameter $\tau_c$ for each cluster, whose formulation follows Eq.~\eqref{eq3}. 


\subsection{Overall Objective} \label{sec4_5}
\vspace{-1.25mm}

In general, instance-wise contrastive learning exploits the local instance-level structures, and prototypical contrastive learning constructs global semantic structures in the latent space. Our approach leverages the advantages of both worlds, and it further injects the semantic constraints from hierarchical prototypes into the learning objective by hierarchical pair selection. Therefore, the overall objective of Hierarchical Contrastive Selective Coding (HCSC) is to learn an image encoder $f_{\theta}$ that minimizes the costs of both instance-wise and prototypical contrastive selective coding: 
\vspace{-2.5mm}
\begin{equation} \label{eq10}
\min \limits_{f_{\theta}} \, \mathcal{L}_{\mathrm{ICSC}} + \mathcal{L}_{\mathrm{PCSC}} .
\end{equation}

\vspace{-1mm}
\section{Experiments} \label{sec_5}
\vspace{-0.75mm}


\subsection{Experimental Setups} \label{sec5_1}
\vspace{-0.5mm}

\textbf{Model details.} The image encoder is with a ResNet-50~\cite{ResNet} backbone, and it maintains another momentum encoder following MoCo v2~\cite{moco_v2}. The implementation of hierarchical K-means (Alg.~\ref{alg:clustering}) is based on faiss~\cite{faiss}, an efficient clustering package. Following SwAV~\cite{SwAV}, we set the number of prototypes at the first hierarchy (\emph{i.e.} $M_1$) as 3000, which fits the statistic patterns of ImageNet~\cite{ImageNet}, \emph{i.e.} our pre-training database. 
We recursively conduct another two rounds of K-means clustering to get two higher-level semantic hierarchies. The clustering hyperparameters are $L=3$ and $(M_1, M_2, M_3) = (3000, 2000, 1000)$, and the sensitivity analysis for $L$ and $\{M_l\}_{l=1}^{L}$ is in Sec.~\ref{sec6_2}. Those clusters with less than 10 samples are discarded to avoid trivial solutions. We directly obtain the assignments $Z_c$ of each cluster from hierarchical K-means results. 
The smooth parameter $\epsilon$ is set as 10 following PCL~\cite{PCL}. Note that, hierarchical prototypes and clustering assignments are fixed during each epoch training, which eases the optimization objective and benefits the convergence of image encoder. 

\textbf{Training details.} We train with an SGD optimizer~\cite{SGD} (weight decay: $1 \times 10^{-4}$; momentum: 0.9; batch size: 256) and a cosine annealing scheduler~\cite{cos} for 200 epochs. The results of longer training are shown in Sec.~\ref{supp_sec4}. 
In the first 20 epochs, the model is warmed up with only the instance-wise loss $\mathcal{L}_{\mathrm{ICSC}}$ (Eq.~\eqref{eq6}). 
The same set of augmentation functions as in MoCo v2~\cite{moco_v2} are used to generate positive pairs, and a queue of 16384 negative candidates are utilized for instance-wise contrastive selective coding (Sec.~\ref{sec4_3}). On 8 Tesla-V100-32GB GPUs, 200 epochs training takes about 4 days \emph{w/o} multi-crop and about 7.5 days \emph{w/} multi-crop.

\textbf{Performance comparisons.} We compare the proposed HCSC approach with superior contrastive learning methods, including NPID~\cite{NPID}, LocalAgg~\cite{LocalAgg}, MoCo~\cite{MOCO}, SimCLR~\cite{SimCLR}, MoCo v2~\cite{moco_v2}, CPC v2~\cite{CPC}, PCL v2~\cite{PCL}, PIC~\cite{PIC}, MoCHi~\cite{MoCHi}, DetCo~\cite{DetCo} and AdCo~\cite{AdCo}. For fair comparison, in the main paper, we report the performance of these methods under 200 epochs training on ImageNet~\cite{ImageNet}. 

\subsection{Experimental Results} \label{sec5_2}



\subsubsection{Linear Classification and KNN Evaluation} \label{sec5_2_1}

\textbf{Evaluation details.} The standard linear classification protocol~\cite{NPID} is adopted, where a linear layer is learned upon the fixed encoder to classify ImageNet~\cite{ImageNet} images. Following PCL~\cite{PCL}, an SGD optimizer (weight decay: 0; momentum: 0.9; batch size: 256) optimizes model for 100 epochs. 

In KNN evaluation~\cite{NPID}, the label of each sample is predicted by aggregating the labels of its nearest neighbors.
Following NPID~\cite{NPID}, we report the highest accuracy of such a KNN classifier over $K \in \{10, 20, 100, 200\}$.

\textbf{Results.} The second column of Tab.~\ref{tab:lincls} compares different approaches on linear classification.
Under the two commonly-used settings with and without multi-crop augmentation~\cite{SwAV}, HCSC outperforms all baseline methods. Especially, HCSC surpasses PCL v2~\cite{PCL} which also incorporates instance-wise and prototypical contrastive learning.
This performance gain illustrates the effectiveness of our pair selection scheme on selecting high-quality positive and negative pairs for instance-wise and prototypical contrast. In Sec.~\ref{supp_sec2}, we provide more linear classification results under different learning configurations. 

For KNN evaluation (the third column of Tab.~\ref{tab:lincls}), when the multi-crop augmentation is not used, HCSC achieves a 3.5\% performance increase compared to previous methods. After adding multi-crop augmentation, HCSC also outperforms the state-of-the-art approach, AdCo~\cite{AdCo}.

\begin{table}[t]
  \caption{Performance comparison on linear and KNN evaluation.} \label{tab:lincls}
  \vspace{-6.5mm}
  \small
  \begin{center}
  \setlength{\tabcolsep}{2.3mm}
  \begin{tabular}{lccc}\hline\hline
    Method & Batch size  & Top1-Acc & KNN-Top1-Acc \\ \hline
    NPID~\cite{NPID} & 256 & 58.5 & 46.8  \\ 
    LocalAgg~\cite{LocalAgg} & 128 & 58.8 & - \\ 
    MoCo~\cite{MOCO}  & 256 & 60.8 & 45.0\dag \\ 
    SimCLR~\cite{SimCLR}  & 256 & 61.9 & 57.4\dag \\ 
    MoCo v2~\cite{moco_v2}  & 256 & 67.5 & 55.8\dag \\ 
    CPC v2~\cite{CPC}  & 512 & 67.6 & - \\ 
    PCL v2~\cite{PCL}  & 256 & 67.6 & 58.1\dag \\ 
    PIC~\cite{PIC} & 512 & 67.6 & 54.7\dag \\ 
    MoCHi~\cite{MoCHi}  & 512 & 67.6 & 57.5\dag  \\ 
    DetCo~\cite{DetCo}  & 256 & 68.6 & 58.9\dag \\
    AdCo~\cite{AdCo}  & 256 & 68.6 & 57.2\dag \\ 
    HCSC  & 256 & \textbf{69.2} & \textbf{60.7} \\ \hline 
    SwAV*~\cite{SwAV}  & 256 & 72.7 & 62.4\dag \\
    AdCo*~\cite{AdCo}  & 256 & 73.2 & 66.3\dag  \\ 
    HCSC*  & 256 & \textbf{73.3} & \textbf{66.6}\\
    \hline
  \end{tabular}
    \end{center}
  \vspace{-1.5mm}
  \footnotesize{* With multi-crop augmentation. \\ \dag$\,$ Evaluated by us with officially released model weights.}
  \vspace{-1.5mm}
\end{table}


\vspace{-2mm}
\subsubsection{Semi-Supervised Learning} \label{sec5_2_2}
\vspace{-0.5mm}

\textbf{Evaluation details.} The encoder and linear classifier are fine-tuned on 1\% or 10\% labeled data in ImageNet.
As in NPID~\cite{NPID}, an SGD optimizer (weight decay: 0; momentum: 0.9; batch size: 256) optimizes model for 70 epochs.

\textbf{Results.} Tab.~\ref{tab:res-semi} presents the results of various methods on semi-supervised learning. Among all approaches, HCSC consistently achieves the best performance under different amount of labeled data.
These results verify that the representations learned by HCSC possess decent global semantic structures that benefit the learning from insufficient data.



\begin{table}[t]
  \caption{Performance comparison on semi-supervised learning.}\label{tab:res-semi}
  \vspace{-6.5mm}
  \small
  \begin{center}
  \setlength{\tabcolsep}{1.5mm}
  \begin{tabular}{lcccc} \hline\hline
  Labeled data & \multicolumn{2}{c}{1\%} & \multicolumn{2}{c}{10\%} \\ \hline
  Method & Top1-Acc & Top5-Acc & Top1-Acc & Top5-Acc  \\
  \hline
  NPID~\cite{NPID} & - & 39.2 & - & 77.4  \\ 
  MoCo v2~\cite{moco_v2}\dag & 36.7 & 64.4 & 60.7 & 83.4 \\
  MoCHi~\cite{MoCHi}\dag & 38.2 & 65.4 & 61.1 & 83.5 \\
  SimCLR~\cite{SimCLR}\dag & 46.8 & 74.2 & 63.6 & \textbf{86.0} \\ 
  PCL v2~\cite{PCL}\dag  & 46.2 & 72.7 & 62.6 & 84.4  \\
  AdCo~\cite{AdCo}\dag & 43.6 & 71.6 & 61.8 & 84.2 \\
  HCSC & \textbf{48.0} & \textbf{75.6} & \textbf{64.3} & \textbf{86.0} \\ 
   \hline
  SwAV*~\cite{SwAV}\dag  & 51.3 & 76.7 & 65.5 & 87.5 \\
  AdCo*~\cite{AdCo}\dag & 54.4 & 79.9 & 66.9 & 87.4 \\
  HCSC*  & \textbf{55.5} & \textbf{80.9} & \textbf{68.7} & \textbf{88.6}\\
  \hline
  \end{tabular}
  \end{center}
  \vspace{-1.5mm}
    \footnotesize{* With multi-crop augmentation. \\ \dag$\,$ Evaluated by us with officially released model weights.}
    \vspace{-1.5mm}
\end{table}


\begin{table}[t]
  \caption{Performance comparison on transfer learning.}\label{tab:res-transfer}
  \vspace{-6.5mm}
  \small
  \begin{center}
  \setlength{\tabcolsep}{2.0mm}
  \begin{tabular}{lcccc}\hline\hline
  Task & \multicolumn{2}{c@{\hspace{3pt}}}{Object Classification} &  \multicolumn{2}{c@{\hspace{12pt}}}{Object Detection} \\ \hline
  Dataset & VOC07 & Places205 & VOC07+12 & COCO \\
  Method  & mAP & Top1-Acc & $\textrm{AP}_{50}$ & AP \\
  \hline
  NPID++~\cite{NPID}  & 76.6 & 46.4 & 79.1 & - \\ 
  MoCo~\cite{MOCO} &  79.2 & 48.9 & 81.1 & - \\
  MoCo v2~\cite{moco_v2} & 84.0 & 50.1 & 82.4 & 40.6\dag \\
  PCL v2~\cite{PCL}  & 85.4 & 50.3 & 78.5 & 41.0\dag \\
  AdCo~\cite{AdCo}\dag & 92.0 & 51.1 & \textbf{82.6} & 41.2 \\
  HCSC  & \textbf{92.8} & \textbf{52.2} & 82.5 & \textbf{41.4} \\
  \hline
  SwAV*~\cite{SwAV}  & 87.6 & 51.2 & 74.5\dag & 38.6\dag \\
  AdCo*~\cite{AdCo}\dag  & 93.1 & 53.9 & 82.7 & 41.4 \\
  HCSC* & \textbf{93.3} & \textbf{55.0} & \textbf{83.2} & \textbf{41.6} \\
  \hline
  \end{tabular}
    \end{center}
    \vspace{-1.5mm}
    \footnotesize{* With multi-crop augmentation. \\ \dag$\,$ Evaluated by us with officially released model weights.}
    \vspace{-2.5mm}
\end{table}


\vspace{-2mm}
\subsubsection{Transfer Learning} \label{sec5_2_3}
\vspace{-0.5mm}

\textbf{Evaluation details.} 
We adopt two classification tasks on PASCAL VOC ~\cite{VOC} and Places205~\cite{Places205} and two detection tasks on PASCAL VOC ~\cite{VOC} and COCO~\cite{COCO} to evaluate transfer learning. The fine-tuning paradigms on these two types of tasks completely follow those in MoCo~\cite{MOCO}. More details on fine-tuning the object detector~\cite{Detectron2} can be found in Sec.~\ref{supp_sec1}. 

\textbf{Results.} Tab.~\ref{tab:res-transfer} presents the comparison among different approaches on transfer learning. Compared to the state-of-the-art method, AdCo~\cite{AdCo}, HCSC obtains better performance on seven of eight experimental settings.
These results demonstrate that the image encoder learned by HCSC succeeds in capturing critical visual patterns shared across different image datasets. In Sec.~\ref{supp_sec3}, we further explore the zero-shot classification on CUB~\cite{CUB}. 




\begin{table}[t]
  \caption{Performance comparison on clustering evaluation.}\label{tab:clustering quality}
  \vspace{-6.5mm}
  \small
  \begin{center}
  \setlength{\tabcolsep}{4.3mm}
  \begin{tabular}{lccc}\hline\hline
  Method  & \# Clusters & NMI & AMI \\ \hline
  DeepCluster~\cite{DeepCluster} & 25000 & - & 0.281 \\
  MoCo v2~\cite{moco_v2} & 25000 & - & 0.285 \\
  PCL v2~\cite{PCL}  & 25000 & 0.616\dag & 0.410 \\
  HCSC  & 25000 & \textbf{0.629} & \textbf{0.462} \\
  \hline
  PCL v2~\cite{PCL}\dag  & 1000 & 0.629 & 0.606 \\
  HCSC & 1000 &\textbf{ 0.638} & \textbf{0.616} \\
  \hline 
  \end{tabular}
  \end{center}
  \vspace{-1.5mm}
    \footnotesize{\dag$\,$ Evaluated by us with officially released model weights.}
\vspace{-1.5mm}
\end{table}


\begin{table}[t]
  \caption{Ablation studies for different components of HCSC. All results are reported on linear classification.}\label{tab:ablation}
  \vspace{-6.5mm}
  \small
  \begin{center}
  \setlength{\tabcolsep}{4mm}
  \begin{tabular}{ccccc|c}\hline\hline
    HP & IL & PL & IS & PS & Top1-Acc \\ \hline
    & \checkmark & \checkmark & & & 67.6 \\
    \checkmark & \checkmark & \checkmark & & & 68.1 \\
    \checkmark & \checkmark & \checkmark  &  & \checkmark &  68.2 \\
    \checkmark & \checkmark & \checkmark & \checkmark & &  68.9 \\
    \checkmark & \checkmark & \checkmark & \checkmark & \checkmark & \textbf{69.2} \\ \hline
    \checkmark & & \checkmark & & \checkmark & 65.7 \\
    \checkmark & \checkmark & & \checkmark & &  68.2 \\
    \hline
  \end{tabular}
\end{center}
  \vspace{-1.5mm}
  \footnotesize{HP: \textbf{H}ierarchical \textbf{P}rototypes; IL: \textbf{I}nstance-wise Contrastive \textbf{L}oss; PL: \textbf{P}rototypical Contrastive \textbf{L}oss; IS: \textbf{I}nstance-wise Pair \textbf{S}election; PS: \textbf{P}rototypical Pair \textbf{S}election}.\\
\vspace{-5mm}
\end{table}


\begin{figure*}[t]
\centering
    \includegraphics[width=1.0\linewidth]{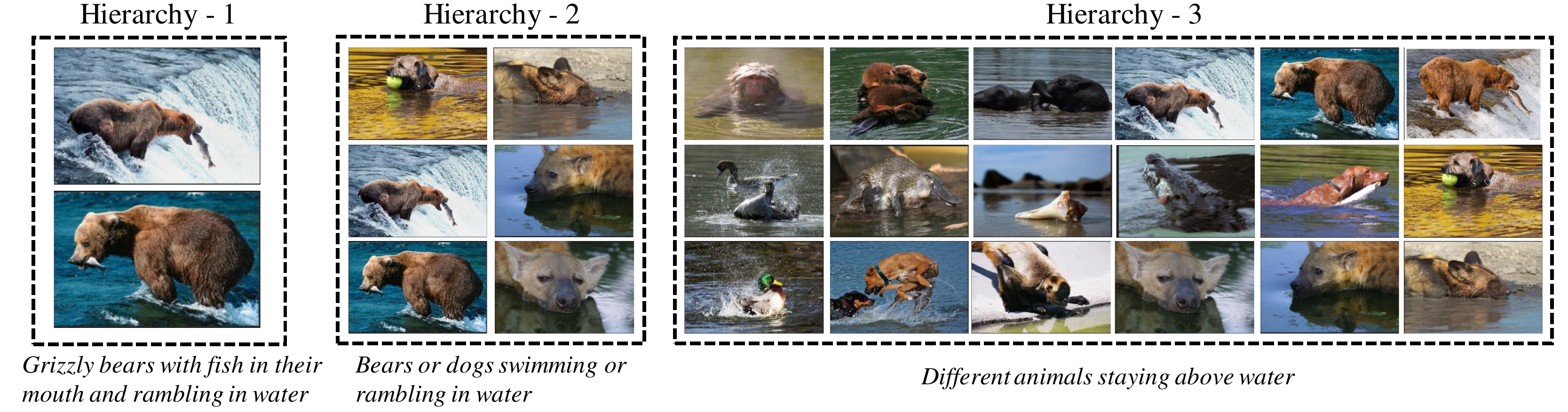}
    \vspace{-6.5mm}
    \caption{Visualize the images associated to a chain of prototypes from the bottom (\emph{i.e.} first) hierarchy to the top (\emph{i.e.} third) hierarchy.}
    \label{fig:clustering visualization}
\vspace{-5mm}
\end{figure*}


\vspace{-2mm}
\subsubsection{Clustering Evaluation} \label{sec5_2_4}
\vspace{-1mm}

\textbf{Evaluation details.} 
We employ two standard metrics for clustering evaluation, \emph{i.e.} Normalized Mutual Information (NMI)~\cite{NMI} and Adjusted Mutual Information (AMI)~\cite{AMI}. Following PCL~\cite{PCL}, the evaluation with 25,000 and 1,000 clusters are respectively performed. More evaluation details are provided in Sec.~\ref{supp_sec1}.  

\textbf{Results.} In Tab.~\ref{tab:clustering quality}, we report the results of clustering evaluation. PCL~\cite{PCL} included MoCo v2~\cite{moco_v2} as a baseline by conducting K-means on the image representations learned by MoCo v2, and we also adopt this baseline in our comparison. Under both the configurations with 25,000 and 1,000 clusters, HCSC clearly surpasses other baseline methods, which illustrates that the semantic hierarchies established in our approach can indeed improve the clustering quality. 






\section{Analysis} \label{sec6}

\subsection{Ablation Study} \label{sec6_1}

\textbf{Effect of hierarchical prototypes.} Under the full model configuration, we fairly compare single prototype hierarchy with multiple prototype hierarchies by using the same number of prototypes for both cases. Comparing the second and fourth row of Tab.~\ref{tab:sensitivity}, we can observe a 1.1\% performance gain after adding two more hierarchies of prototypes, which verifies the benefit of using hierarchical prototypes.

 
\textbf{Effect of pair selection.} In Tab.~\ref{tab:ablation}, a 0.8\% performance gain is obtained after adding the negative pair selection scheme in instance-wise contrastive learning (fourth row \emph{vs.} second row). Prototypical pair selection is less effective when applied individually (third row). However, by combining two pair selection schemes, the full model achieves the highest accuracy of 69.2\% (fifth row), which demonstrates the complementarity of two pair selection schemes. 

\textbf{Effect of instance-wise and prototypical contrastive loss.} From the last three rows of Tab.~\ref{tab:ablation}, we observe a 3.5\% (1.0\%) performance decay when the instance-wise (prototypical) contrastive loss is removed, which proves that these two kinds of contrastive losses are complementary.


\begin{table}[t]
  \caption{Sensitivity analysis on the number of hierarchies and the number of prototypes. All results are on linear classification.}\label{tab:sensitivity}
  \vspace{-6.5mm}
  \small
  \begin{center}
  \setlength{\tabcolsep}{5.3mm}
  \begin{tabular}{lc}\hline\hline
    Configuration of prototypes & Top1-Acc \\ \hline
    3000 & 68.0 \\
    6000 & 68.1 \\
    3000-2000 & 69.0 \\
    3000-2000-1000 & 69.2 \\
    3000-2000-1000-500 & 69.2 \\
    \hline
    1000-500-200 & 68.7 \\
    3000-2000-1000 & 69.2 \\
    10000-5000-1000 & 69.2 \\
    30000-10000-1000 & 69.3 \\
    \hline
  \end{tabular}
  \end{center}
  \vspace{-5mm}
\end{table}

\subsection{Sensitivity Analysis} \label{sec6_2}

\textbf{Sensitivity to the number of semantic hierarchies.} In the first section of Tab.~\ref{tab:sensitivity}, 
we find that more semantic hierarchies (3 or 4 hierarchies) obviously benefit model's performance on linear classification. Under such settings, the hierarchical semantic structures underlying the pre-training database can be well captured by hierarchical prototypes.

\textbf{Sensitivity to the number of prototypes.} According to the second section of Tab.~\ref{tab:sensitivity},
the three configurations with sufficient prototypes (second to fourth rows) perform comparably well, while the configuration with insufficient prototypes (first row) performs worse. These results illustrate the importance of using abundant prototypes to fully capture the semantic clusters underlying the data.  




\begin{figure}[t]
\centering
    \includegraphics[width=0.85\linewidth]{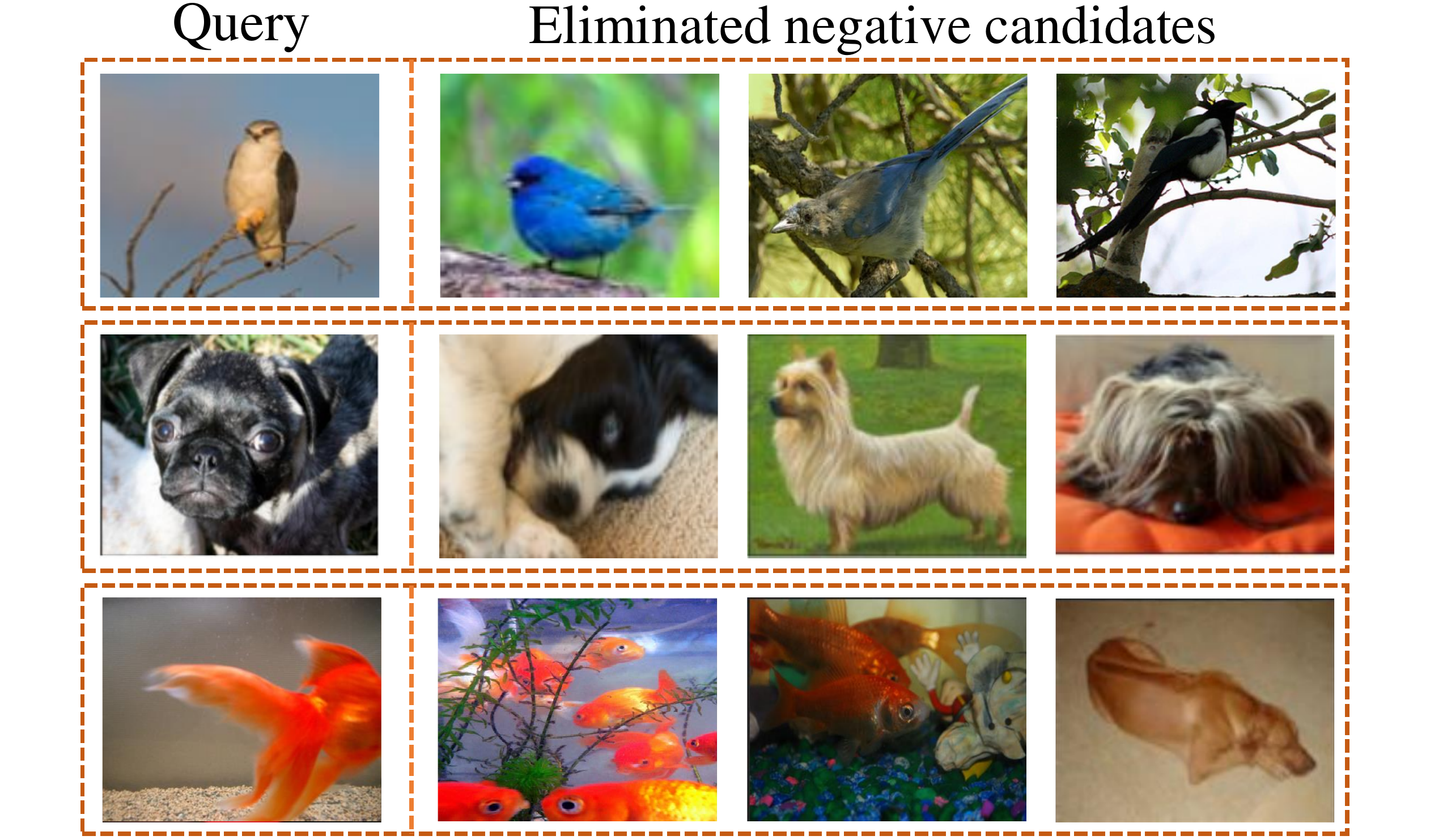}
    \vspace{-2mm}
    \caption{Visualize the query sample and the negative candidates eliminated by our pair selection approach.}
    \label{fig:mining visualization}
\vspace{-3.5mm}
\end{figure}


\subsection{Visualization} \label{sec6_3}

\textbf{Visualization of hierarchical semantic structures.} In Fig.~\ref{fig:clustering visualization}, we visualize the images assigned to a chain of prototypes,
which capture some interesting semantic hierarchies, \emph{i.e.} ``grizzly bears with fish in their mouth and rambling in water'' $\rightarrow$ ``bears or dogs swimming or rambling in water'' $\rightarrow$ ``different animals staying above water''. More visualization results can be found in Sec.~\ref{supp_sec7}. 


\textbf{Visualization of pair selection.} Fig.~\ref{fig:mining visualization} shows the query sample and the negative candidates eliminated by our selection scheme.
Most of these eliminated candidates own similar semantics with the query, which verifies the effectiveness of the proposed instance-wise pair selection.





\vspace{-0.5mm}
\section{Conclusions and Future Work} \label{sec7}
\vspace{-0.5mm}

This work proposes a novel contrastive learning framework, Hierarchical Contrastive Selective Coding (HCSC). In this framework, the hierarchical semantic structures underlying the data are captured by hierarchical prototypes. Upon these prototypes, a novel pair selection scheme is designed to better select positive and negative pairs for contrastive learning. 
Extensive experiments on various downstream tasks verify the superiority of our HCSC method. 

The main limitation of the current HCSC method is that the hierarchical prototypes discovered during pre-training are discarded in downstream tasks. However, these prototypes contain rich semantic information, and they should benefit the semantic understanding of a downstream application in some way. Therefore, our future work will mainly focus on enhancing model's performance on downstream tasks by fully utilizing hierarchical prototypes.



\section{Acknowledgement} \label{sec8}

Yi Xu is supported by National Natural Science Foundation of China 62171282, 111 project BP0719010, STCSM 18DZ2270700, Key Research and Development Program of Chongqing (cstc2021jscx-gksbX0032), and Shanghai Municipal Science and Technology Major Project 2021SHZDZX0102. Bingbing Ni is supported by National Science Foundation of China (U20B2072, 61976137). Authors would like to thank ByteDance for providing GPUs. Authors also appreciate Jian Zhang, Jie Zhou and Hannes St{\"a}rk for the inspiring discussions.




{\small
\bibliographystyle{ieee_fullname}
\bibliography{egbib}
}
\clearpage
\appendix

\section{More Implementation Details} \label{supp_sec1}

\textbf{Linear classification.} For performance comparison, we follow the learning configuration of PCL~\cite{PCL} to train the linear classifier with an SGD optimizer (weight decay: 0; momentum: 0.9; batch size: 256) for 100 epochs. The learning rate is initialized as 5.0 and decayed by a factor of 0.1 at the 60th and 80th epoch.

\textbf{KNN evaluation.} We follow NPID~\cite{NPID} to design a KNN classifier which predicts the label of each sample by aggregating the labels of its nearest neighbors. Specifically, given a test image $x$, we first extract its embedding $z$ using the pre-trained encoder. This embedding vector is compared against the embeddings of all other images in the dataset, and a cosine similarity score $s_{\mathrm{cos}}(z, z_i)$ is computed for each image pair. According to these similarity scores, we select the top K nearest neighbors of the test image, denoted as $\mathcal{N}_K(x)$. On such basis, we compute the unnormalized likelihood $p_c(x)$ that the test image belongs to class $c$ via a weighted voting:
\begin{equation} \label{supp_eq1}
p_c(x) = \sum_{x_i \in \mathcal{N}_K(x)} \mathbbm{1}(y_i = c) \exp \big( s_{\mathrm{cos}}(z, z_i) / \tau_{\mathrm{KNN}} \big) ,
\end{equation}
where $\mathbbm{1}(y_i = c)$ is an indicator function judging whether the sample $x_i$ belongs to class $c$, and the temperature parameter $\tau_{\mathrm{KNN}}$ is set as 0.07 following NPID. Based on these likelihoods, the KNN classifier predicts the category of $x$ as $y = \mathop{\arg\max}_{c \in C} p_c(x)$. As in NPID, the final result of KNN evaluation is reported as the highest classification accuracy over $K \in \{10, 20, 100, 200\}$.

\textbf{Semi-supervised learning.} In this experiment, we follow NPID~\cite{NPID} to fine-tune the image encoder and linear classifier with an SGD optimizer (weight decay: 0; momentum: 0.9; batch size: 256) for 70 epochs. The learning rate is initialized as 0.005 and decayed by a factor of 0.1 at the 30th and 60th epoch.

\textbf{Transfer learning.} This experiment involves two types of transfer learning tasks, \emph{i.e.} object classification and object~detection. We strictly follow the fine-tuning paradigms of MoCo~\cite{MOCO} on these two types of tasks.

For object classification, our model is evaluated on PASCAL VOC~\cite{VOC} and Places205~\cite{Places205} datasets. We follow the standard dataset splits of VOC07 and Places205 to perform training and testing. On both datasets, following SwAV~\cite{SwAV}, we keep the pre-trained encoder fixed and learn a linear layer for classification. On PASCAL VOC, the linear classifier is trained for 100 epochs by an SGD optimizer (weight decay: 0; momentum: 0.9; batch size: 16), and the initial learning rate of 0.05 is adjusted by a cosine annealing scheduler~\cite{cos}. On Places205, we train the linear classifier with an SGD optimizer (weight decay: 0; momentum: 0.9; batch size: 256) for 100 epochs, and the initial learning rate of 3.0 is adjusted by a cosine annealing scheduler.

For object detection, we evaluate our model on PASCAL VOC~\cite{VOC} and COCO~\cite{COCO} datasets. On PASCAL VOC, the training and validation splits of VOC07+12 is used for training, and the test split of VOC07 is used for evaluation. Faster-RCNN-C4~\cite{faster_rcnn} serves as the object detector. We initialize its ResNet-50 backbone with the weights pre-trained by our HCSC approach, and the whole detection model is fine-tuned for 24,000 iterations by an SGD optimizer (weight decay: $1 \times 10^{-4}$; momentum: 0.9; batch size: 16). The initial learning rate of 0.02 is warmed up for 100 iterations and decayed by a factor of 0.1 at the 18,000th and 22,000th iteration. On COCO, the detection model is trained on the train2017 subset for 180,000 iterations, and it is then evaluated on the val2017 subset. An identical SGD optimizer as in PASCAL VOC experiment is employed, and the initial learning rate of 0.02 is warmed up for 100 iterations and decayed by a factor of 0.1 at the 120,000th and 160,000th iteration.

\textbf{Clustering evaluation.} Following PCL~\cite{PCL}, the clustering evaluation with 25,000 and 1,000 clusters are respectively performed. For the experiment using 25,000 clusters, we train an HCSC model with three prototype hierarchies 25000-10000-1000, and the bottom hierarchy with 25,000 prototypes are used for evaluation. For the experiment using 1,000 clusters, an HCSC model with three prototype hierarchies 3000-2000-1000 is trained, and we utilize the top hierarchy with 1,000 prototypes for evaluation.  




\begin{table}[t]
  \caption{Performance comparison on linear classification under different learning configurations.}
  \label{tab:lincls configs}
  \vspace{-6.5mm}
  \small
  \begin{center}
   \setlength{\tabcolsep}{0.58mm}
  \begin{tabular}{lccccc}
    \hline\hline
    Method & Config & Initial lr & Scheduler & Top1-Acc \\ \hline
    PCL v2~\cite{PCL} & PCL~\cite{PCL} & 5.0 & step(0.1, [60,80]) & 67.6 \\
    HCSC & PCL~\cite{PCL} & 5.0 & step(0.1, [60,80]) & \textbf{69.2} \\
    \hline
    AdCo~\cite{AdCo} & AdCo~\cite{AdCo} & 10.0 & cosine & 68.6 \\
    HCSC & AdCo~\cite{AdCo} & 10.0 & cosine & \textbf{68.9} \\
    \hline
    MoCo v2~\cite{moco_v2} & MoCo v2~\cite{moco_v2} & 30.0 & step(0.1, [60,80]) & \textbf{67.5} \\
    HCSC & MoCo v2~\cite{moco_v2} & 30.0 &  step(0.1, [60,80]) & 67.3 \\
    \hline
  \end{tabular}
  \end{center}
  \vspace{-4.5mm}
\end{table}


\section{More Results of Linear Classification} \label{supp_sec2}

We notice that the fine-tuning configuration vary across previous works when performing linear classification on ImageNet~\cite{ImageNet}. Therefore, in Tab.~\ref{tab:lincls configs}, we further evaluate our HCSC model under the configurations from three different works, \emph{i.e.} PCL~\cite{PCL}, AdCo~\cite{AdCo} and MoCo v2~\cite{moco_v2}. Under the learning configurations of PCL and AdCo, the performance difference of HCSC is merely 0.3\%, and it outperforms these two approaches on their respective configurations. These results verify the robustness of our method when varying the initial learning rate between 5.0 and 10.0 and changing between a step scheduler decaying twice and a cosine annealing scheduler. On the configuration of MoCo v2, HCSC suffers an obvious performance decrease and performs worse than MoCo v2. This negative result illustrates that too high initial learning rate, like 30.0 in MoCo v2's configuration, will hamper the effectiveness of HCSC during downstream fine-tuning. 



\section{Zero-Shot Classification on CUB} \label{supp_sec3}

In this section, we study a more difficult transfer learning problem, \emph{i.e.} directly transferring the encoder learned on ImageNet~\cite{ImageNet} to a fine-grained classification dataset, Caltech-UCSD-Birds (CUB)~\cite{CUB}, without learning a task-specific classifier. Therefore, this problem can be regarded as a \textbf{cross-domain zero-shot classification} problem, and it evaluates whether a self-supervised learning method can capture fine-grained semantic structures by pre-training on a general-purpose database, like ImageNet. 

\textbf{Evaluation details.} We evaluate model's zero-shot classification performance on CUB with the standard KNN evaluation protocol. Specifically, a KNN classifier is employed to predict the label of each sample by aggregating the labels of its nearest neighbors. The implementation details of such a KNN classifier is specified in the KNN evaluation part of Sec.~\ref{supp_sec1}. We report the highest accuracy of the KNN classifiers over $K \in \{10, 20, 100, 200\}$, which follows NPID~\cite{NPID}.

\textbf{Results.} Tab.~\ref{tab:knn-cub} presents the performance comparison among different approaches on this task. Under both the configurations with and without multi-crop augmentation, HCSC clearly outperforms other baseline methods. This superior performance demonstrates that, by pre-training with HCSC, the image encoder can well capture the fine-grained semantic structures underlying an image dataset, and such a capability can even be transferred to other datasets.  





\begin{table}[t]
  \caption{Performance comparison on zero-shot classification. This experiment transfers the encoder learned on ImageNet to CUB.}\label{tab:knn-cub}
  \vspace{-6.5mm}
  \small
  \begin{center}
  \setlength{\tabcolsep}{6.5mm}
  \begin{tabular}{lc}\hline\hline
    Method & KNN-Top1-Acc \\ \hline
    MoCo~\cite{MOCO}\dag & 19.5 \\
    MoCo v2~\cite{moco_v2}\dag & 23.1  \\
    SimCLR~\cite{SimCLR}\dag & 23.9 \\
    PIC~\cite{PIC}\dag & 18.2 \\
    PCL v2~\cite{PCL}\dag & 22.3  \\
    AdCo~\cite{AdCo}\dag & 22.9   \\
    HCSC & \textbf{26.9} \\
    \hline
    SwAV*~\cite{SwAV}\dag & 26.2  \\
    AdCo*~\cite{AdCo}\dag  & 30.6  \\
    HCSC*  & \textbf{31.5}  \\
    \hline
  \end{tabular}
  \end{center}
  \vspace{-1.5mm}
  \footnotesize{* With multi-crop augmentation. \\ \dag$\,$ Evaluated by us with officially released model weights.}
  \vspace{-1mm}
\end{table}


\begin{table}[t]
  \caption{Performance of models under different training epochs. The results are reported on linear and KNN evaluation.} \label{tab:model_zoo_epochs}
  \vspace{-6.5mm}
  \small
  \begin{center}
  \setlength{\tabcolsep}{0.6mm}
  \begin{tabular}{lcccc}\hline\hline
    Method & Epochs & Batch size & Top1-Acc & KNN-Top1-Acc \\ 
    \hline
    NPID~\cite{NPID} & 200 & 256 & 58.5 & 46.8  \\ 
    LocalAgg~\cite{LocalAgg} & 200 & 128 & 58.8 & - \\ 
    MoCo~\cite{MOCO} & 200 & 256 & 60.8 & 45.0\dag \\ 
    SimCLR~\cite{SimCLR}  & 200 & 256 & 61.9 & 57.4\dag \\ 
    MoCo v2~\cite{moco_v2}  & 200 & 256 & 67.5 & 55.8\dag \\ 
    CPC v2~\cite{CPC}  & 200 & 512 & 67.6 & - \\ 
    PCL v2~\cite{PCL}  & 200 & 256 & 67.6 & 58.1\dag \\ 
    PIC~\cite{PIC} & 200 & 512 & 67.6 & 54.7\dag \\ 
    MoCHi~\cite{MoCHi}  & 200 & 512 & 67.6 & 57.5\dag  \\ 
    DetCo~\cite{DetCo}  & 200 & 256 & 68.6 & 58.9\dag \\
    AdCo~\cite{AdCo}  & 200 & 256 & 68.6 & 57.2\dag \\ 
    HCSC  & 200 & 256 & \textbf{69.2} & \textbf{60.7} \\
    \hline
    SwAV*~\cite{SwAV}  & 200 & 256 & 72.7 & 62.4\dag \\
    AdCo*~\cite{AdCo}  & 200 & 256 & 73.2 & 66.3\dag  \\ 
    HCSC*  & 200 & 256 & \textbf{73.3} & \textbf{66.6}\\
    \hline
    DeepCluster-v2~\cite{SwAV} & 400 & 4096 & 70.2 & 62.4\dag \\
    SeLa-v2~\cite{SwAV} & 400 & 4096 & 67.2 & 57.9\dag \\
    SwAV~\cite{SwAV} & 400 & 4096 & 70.1 & 61.3\dag \\
    HCSC & 400 & 256 & \textbf{71.0} & \textbf{64.1} \\
    \hline
    DeepCluster-v2*~\cite{SwAV} & 400 & 4096 & 74.3 & 66.0\dag \\
    SeLa-v2*~\cite{SwAV} & 400 & 4096 & 71.8 & 61.7\dag \\
    SwAV*~\cite{SwAV} & 400 & 256 & 74.3 & 64.3\dag \\
    SwAV*~\cite{SwAV} & 400 & 4096 & \textbf{74.6} & 65.0\dag \\
    HCSC* & 400 & 256 & 74.1 & \textbf{69.9} \\
    \hline
    MoCo v2~\cite{moco_v2} & 800 & 256 & 71.1 & 61.8\dag \\
    HCSC & 800 & 256 & \textbf{72.0} & \textbf{64.5} \\
    \hline
    DeepCluster-v2*~\cite{SwAV} & 800 & 4096 & 75.2 & 66.9\dag \\
    SwAV*~\cite{SwAV} & 800 & 4096 & \textbf{75.3} & 65.7\dag \\
    HCSC* & 800 & 256 & 74.2 & \textbf{70.6} \\
    \hline
  \end{tabular}
    \end{center}
  \vspace{-1.5mm}
  \footnotesize{* With multi-crop augmentation. \\ \dag$\,$ Evaluated by us with officially released model weights.}
  \vspace{-1.5mm}
\end{table}


\section{Model Zoo} \label{supp_sec4}

To make this project a more solid contribution, we train a comprehensive set of models, including longer training epochs, single- and multi-crop settings and more backbone architectures, and we will continually release corresponding codes and model weights to the community. 

\subsection{Models of Longer Training} \label{supp_sec4_1}

In Tab.~\ref{tab:model_zoo_epochs}, we give comprehensive comparisons among various methods under different training epochs. The results show that, under 400 and 800 epochs training, our HCSC method obviously outperforms previous state-of-the-art approaches on KNN evaluation, and its linear evaluation performance is competitive.


\subsection{Models with Different Architectures} \label{supp_sec4_2}

This part of works are in progress.



\section{Time Complexity Analysis} \label{supp_sec5}

HCSC involves an extra hierarchical K-means step for each epoch. Here, we compare it with another clustering-based method, SwAV~\cite{SwAV}. In each training step, SwAV performs three iterations of Sinkhorn-Knopp algorithm to update clustering assignments, which has a time complexity of $\mathcal{O}(KM)$ ($K$: batch size; $M$: number of prototypes). After amortizing the cost of hierarchical K-means to all training steps within an epoch, our HCSC method has an extra time complexity of $\mathcal{O}(NM_1 + M_1M_2 + M_2M_3) / T = \mathcal{O}(KM_1)$ for each step ($N$: dataset size; $M_l$: number of prototypes at the $l$-th hierarchy; $T$: training steps per epoch). Therefore, when it holds that $M \approx M_1$, SwAV and HCSC have comparable extra computation. In the first two rows of Tab.~\ref{tab:time_complexity}, we compare the per-epoch running time of SwAV (with 3000 prototypes) and the vanilla HCSC (with 3000-2000-1000 hierarchical prototypes), which makes $M = M_1$. The comparable cost of time supports the analysis above.

To further enhance the efficiency of HCSC, we employ faiss~\cite{faiss}, a library for efficient similarity search and clustering, to perform the hierarchical K-means step. Thanks for the high parallelism of faiss, the improved HCSC model, \emph{i.e.} HCSC (parallel), achieves much better computational efficiency than the vanilla HCSC, \emph{i.e.} HCSC (non-parallel), as shown in the last two rows of Tab.~\ref{tab:time_complexity}.



\begin{table}[t]
  \caption{Per-epoch running time comparison (batch size: 256).}\label{tab:time_complexity}
  \vspace{-6.5mm}
  \small
  \begin{center}
  \setlength{\tabcolsep}{3.1mm}
  \begin{tabular}{c|c|c}
  \hline \hline
  Method & \emph{w/o} multi-crop & \emph{w/} multi-crop \\
  \hline
  SwAV~\cite{SwAV} & 27min 53s & 44min 30s \\
  HCSC (non-parallel) & 26min 22s & 44min 59s \\
  HCSC (parallel) & \textbf{21min 11s} & \textbf{39min 22s} \\
  \hline 
  \end{tabular}
  \end{center}
\vspace{-4mm}
\end{table}


\section{Analysis on Contrastive Selective Coding} \label{supp_sec6}

Here, we analyze the proposed instance-wise and prototypical contrastive selective coding from two perspectives: (1) how it can select more diverse positive pairs with similar semantics, and (2) how it can select more precise negative pairs with truly distinct semantics. Though the pre-training stage is unsupervised, the labels and label hierarchies of the pre-training database, ImageNet~\cite{ImageNet}, are publicly available to enable us to perform this analysis.

\subsection{Analysis on Positive Pair Selection} \label{supp_sec6_1}

In this study, we aim to verify that our method can better include \textbf{images and their corresponding prototypes at higher ImageNet label hierarchy} as positive pairs. In Tab.~\ref{tab:label_ami}, we report the adjusted mutual information (AMI) between prototypes and the ImageNet labels at three hierarchies. Compared with the prototypes with a single hierarchy, the prototypes with three hierarchies can better capture the semantics on all three label hierarchies. Hence, the positive image-prototype pairs selected based on our hierarchical prototypes are more semantically diverse.

\subsection{Analysis on Negative Sample Selection} \label{supp_sec6_2}

\begin{figure}[t]
\centering
    \includegraphics[width=1.0\linewidth]{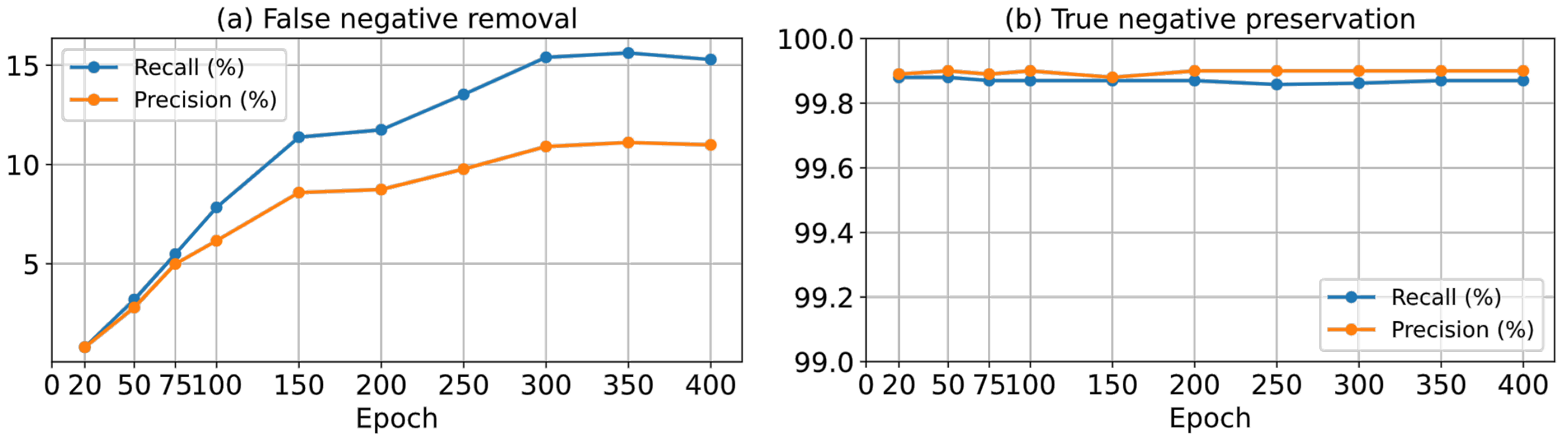}
    \vspace{-7mm}
    \caption{Performance of our negative sample selection scheme.}
    \label{fig:negative_filtering}
\vspace{-1.5mm}
\end{figure}

This study seeks to measure the effectiveness of our negative sample selection scheme. In Fig.~\ref{fig:negative_filtering}, we plot the precision and recall of false negatives and true negatives along training. This recording shows \textbf{stably growing false negative removal} and \textbf{constantly high true negative preservation}, which verifies that the proposed scheme can keep most of the correct negative samples and, at the same time, eliminate more and more false negatives as the representation quality improves. 


\begin{table}[t]
  \caption{Adjusted Mutual Information (AMI) between prototypes and ImageNet labels on the 1st, 2nd and 3rd label hierarchy (count from bottom to top).}\label{tab:label_ami}
  \vspace{-6.5mm}
  \small
  \begin{center}
  \setlength{\tabcolsep}{1.33mm}
  \begin{tabular}{c|ccc}
  \hline \hline
  Prototype Config & 1st hierarchy & 2nd hierarchy & 3rd hierarchy \\
  \hline
  6000 & 0.543 & 0.535 & 0.506 \\
  3000-2000-1000 & \textbf{0.582} & \textbf{0.588} & \textbf{0.566} \\
  \hline 
  \end{tabular}
  \end{center}
\vspace{-3mm}
\end{table}


\section{More Visualization Results} \label{supp_sec7}

\subsection{Visualization of Hierarchical Semantics} \label{supp_sec7_1}

\begin{figure*}[t]
\centering
    \includegraphics[width=1.0\linewidth]{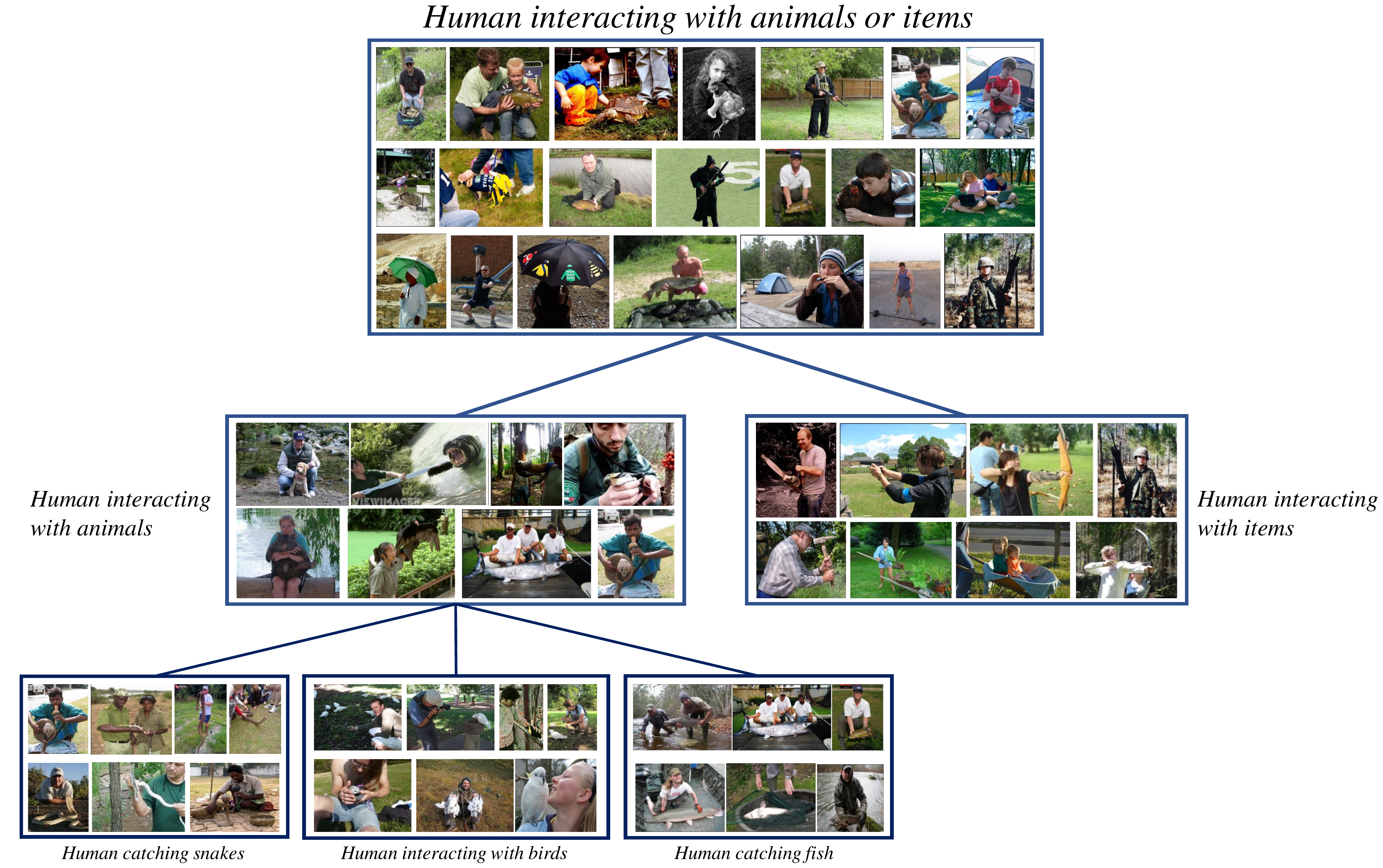}
    \vspace{-5mm}
    \caption{Visualization of a typical substructure of hierarchical prototypes.}
    \label{fig:clustering tree}
\end{figure*}

In Fig.~\ref{fig:clustering tree}, we visualize the images assigned to the prototypes in a substructure of hierarchical prototypes. The semantics of the images assigned to the prototype at top hierarchy are most diverse, which represents the coarse-grained semantics of ``human interacting with animals or items''. By comparison, the images assigned to the prototypes at bottom hierarchy express finer-grained semantics, \emph{e.g.} ``human catching snakes'', ``human interacting with birds'' and ``human catching fish''. These results illustrate that the proposed hierarchical prototypes can indeed capture hierarchical semantic structures. 

\subsection{Visualization of Feature Representations} \label{supp_sec7_2} 

\begin{figure*}[t]
\centering
    \includegraphics[width=1.0\linewidth]{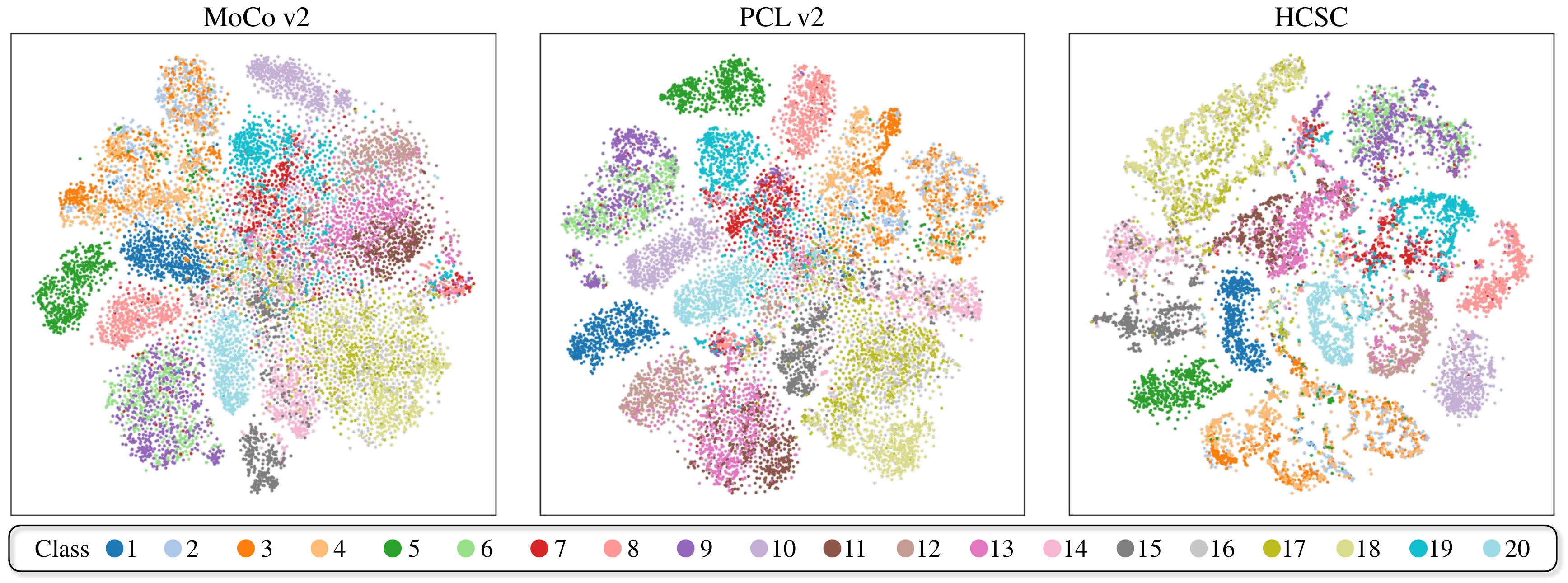}
    \vspace{-5mm}
    \caption{The t-SNE visualization of the learned representations for ImageNet training samples from the first 20 classes.}
    \label{fig:tsne visualization}
\end{figure*}

In Fig.~\ref{fig:tsne visualization}, we use t-SNE~\cite{tsne} to visualize the representations of ImageNet~\cite{ImageNet} images learned by three methods, \emph{i.e.} MoCo v2~\cite{moco_v2}, PCL v2~\cite{PCL} and the proposed HCSC, in which the first 20 classes of ImageNet are visualized following PCL~\cite{PCL}. The image representations learned by MoCo v2 are not separable among many classes. By comparison, PCL v2 derives more separable representations among different classes, while it confuses the image representations of class 7, 19 and 20. HCSC produces more separable feature representations among these three classes, and the representations from all 20 classes are best separated under our approach. These visualization results demonstrate that HCSC can derive discriminative feature representations which benefit various downstream tasks. 



\end{document}